%% file: 0-root.tex
\documentclass[journal]{IEEEtran}
\IEEEoverridecommandlockouts
\usepackage{lipsum}
\usepackage{hyperref}
\usepackage{placeins}
\usepackage{mathtools}
\usepackage{graphicx}
\usepackage{float}
\usepackage{setspace}
\usepackage{color}
\usepackage[dvipsnames]{xcolor}
\usepackage{cite}
\usepackage{indentfirst}
\usepackage{amssymb}
\usepackage{amsmath}
\usepackage{amsbsy}
\usepackage{subfigure}
\usepackage{graphicx}
\usepackage{setspace}
\usepackage{soul}
\usepackage[ruled,vlined]{algorithm2e}

\begin{document}

\title{Vision-based Navigation for a Small-scale Quadruped Robot Pegasus-Mini}

\author{Ganyu Deng\textsuperscript{$1,\dagger$}, Jianwen Luo\textsuperscript{$1,2,\dagger$}, Caiming Sun\textsuperscript{$1,2,*$}, Dongwei Pan\textsuperscript{1}, Longyao Peng\textsuperscript{1}, Ning Ding\textsuperscript{$1,2$}, Aidong Zhang\textsuperscript{1,2}
\thanks{This work was supported in part by National Natural Science Foundation of China under Grant 51905251 and Grant U1613223; in part by Shenzhen Institute of Artificial Intelligence and Robotics for Society (AIRS) project under Grant AC01202101023. \textit{($*$Corresponding author: Caiming Sun. Email: cmsun@cuhk.edu.cn)}}
\thanks{$^{1}$Shenzhen Institute of Artificial Intelligence and Robotics for Society (AIRS), The Chinese University of Hong Kong, Shenzhen, Shenzhen, 518172, China.}
\thanks{$^{2}$Institute of Robotics and Intelligent Manufacturing (IRIM), The Chinese University of Hong Kong, Shenzhen, Shenzhen, 518172, China.}
}

\markboth{Manuscript has been accepted by IEEE ROBIO 2021}%
{Shell \MakeLowercase{\textit{et al.}}: The manuscript title}

\maketitle

\begingroup\renewcommand\thefootnote{$\dagger$}
\footnotetext{Ganyu Deng and Jianwen Luo are co-first authors.}
\endgroup

\begin{abstract}
\input{1-abstact}
\end{abstract}


\input{2-intro}
\vspace{-3mm}
\input{3-related_work}
\vspace{-3mm}
\input{4-system_overview}
\input{5-semantic_segmentation}
\vspace{-3 mm}
\input{6-path_planning}
\vspace{-7 mm}
\input{7-experiment}
\vspace{-5 mm}
\input{8-conclusion}

\section*{Acknowledgment}
All authors would like to thank Ms. Jing Lin, Mr. Wu Shi, and Zuwen Zhu for the design and implementation of the Pegasus-Mini.

\bibliographystyle{IEEEtran}
\bibliography{9-mybib}

\end{document}

%% file: 1-abstact.tex
Quadruped locomotion is currently a vibrant research area, which has reached a level of maturity and performance that enables some of the most advanced real-world applications with autonomous quadruped robots both in academia and industry. Blind robust quadruped locomotion has been pushed forward in control and technology aspects within recent decades. However, in the complicated environment, the capability including terrain perception and path planning is still required. Visual perception is an indispensable ability in legged locomotion for such a demand. This study explores a vision-based navigation method for a small-scale quadruped robot Pegasus-Mini, aiming to propose a method that enables efficient and reliable navigation for the small-scale quadruped locomotion. The vision-based navigation method proposed in this study is applicable in such a small-scale quadruped robot platform in which the computation resources and space are limited. The semantic segmentation based on a CNN model is adopted for the real-time path segmentation in the outdoor environment. The desired traverse trajectory is generated through real-time updating the middle line, which is calculated from the edge position of the segmented path in the images. To enhance the stability of the path planning directly based on the semantic segmentation method, a trajectory compensation method is supplemented considering the temporal information to revise the untrustworthy planned path. Experiments of semantic segmentation and navigation in a garden scene are demonstrated to verify the effectiveness of the proposed method.

%% file: 2-intro.tex
\section{Introduction}

\IEEEPARstart{Q}{uadruped} locomotion has been a vibrant research field within recent years. Although blindly legged locomtion is extensively studied \cite{rudin2021cat, kim2020dynamic, 8324642, luo2017advanced}, autonomous navigation beyond blind robust locomotion in the complicated outdoor environment has drawn more interest. To accomplish the outdoor navigation, it is required to actively perceive the environment and efficiently plan the locomotion path accordingly. Currently, autonomous navigation for quadruped locomotion is mainly focused on medium-scale or large-scale real robot platforms, which are able to offer relatively larger load capacity and space for high-performance computing equipment which is critical for the processing and computation of big data from the perception sensors such as Lidar, camera, etc. 

One of the paradigm is the ANYmal series, a type of large-scale quadruped robot, which weighs around 30 $kg$. Its length, height and width are around 1 $m$, 0.8 $m$ and 0.5 $m$ respectively. The earliest ANYmal is equipped with a rotating Hokuyo UTM-30lx laser sensor for navigation and three intel NUC PC for computing \cite{7758092}. A traversability map is built for the path planning. RRT algorithm is adopted to optimize the path length and safety \cite{7759199}. A method is presented on ANYmal, which rebuilds the surrounding environment in form of an elevation map with a depth camera \cite{8392399}. This approach updates the probabilistic map step by step. In another implementation, ANYmal precisely quantifies the environment traversability with the consideration of various geometry factors like surface roughness, inclination, and height difference \cite{7759199}. These methods show impressive performance and high precision which brings convenience for the foothold selection and path planning. However, efficiency is relatively low in the implementation due to the time-consuming 3D reconstruction process. The navigation research on ANYmal mainly focuses on the traditional approaches, such as geometry feature, point cloud, elevation map, or occupancy grid map. 

\begin{figure}[t]
\centering
\includegraphics[width = 3.4 in]{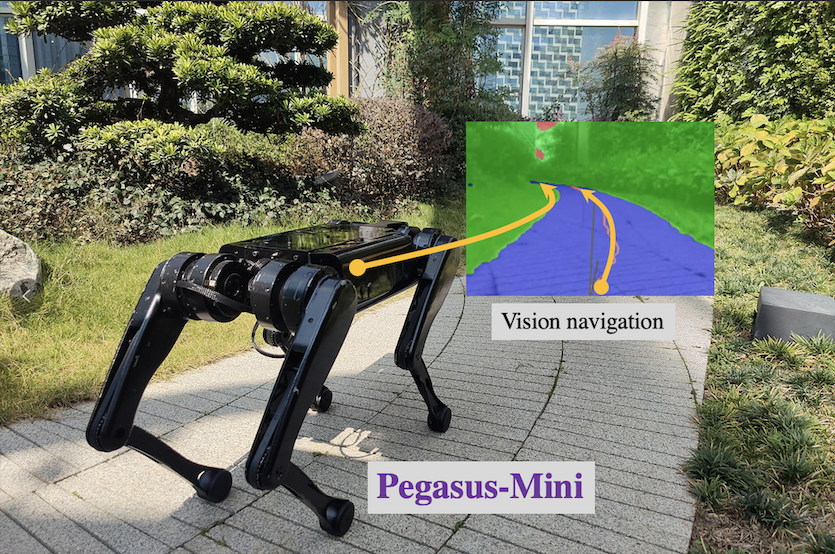}
\caption{Locomotion of a small\textcolor{blue}{-}scale quadruped robot, Pegasus-Mini, in a garden with the assistance of vision-based navigation.}
\label{fig:concept}
\end{figure}

Another well-known quadruped robot is HyQ, which weighs 85 $kg$ and measures approximate 1.0$\times$0.5$\times$0.98 $m$ (length$\times$width$\times$height) \cite{9133154}. A depth camera (Asus Xtion), a MultiSense SL sensor (2.6 $kg$), and two intel i5 processors are mounted on HyQ to realize a coupled framework consisting of motion planning, whole-body control, and terrain model. A real-time and dynamic foothold adaptation strategy based on visual feedback is also presented on HyQ \cite{8642374}. Similar work also includes \cite{8961842}, which improve the energy efficiency based on the perception of the environment. These quadruped robots are large-scale with higher load capacity and therefore are able to integrate a variety of navigation sensors and high-performance computing platforms. Real-time path planning for navigation based on images is not fully explored yet.

Among small-scale quadruped robots, MIT Mini-cheetah, as one of the canonical platforms, presents the application in obstacle avoidance during the navigation \cite{9196777}. MIT Mini-cheetah is 0.3 $m$ tall and 9 $kg$. The small body size limits the types and numbers of sensors and hence computing performance is discounted. Currently, only two cameras and a depth camera are mounted on MIT Mini-cheetah. 

Compared with the existing research in quadruped navigation, vision-based navigation, especially using semantic segmentation, for real-time path planning on a small-scale quadruped robot platform is yet explored.

The contributions in this letter lie in the following twofold:

1) Implementation of a vision-based navigation using semantic segmentation on a lightweight computing architecture deployed on a small-scale quadruped robot.

2) Trajectory compensation method is proposed to enhance the success rate of the vision-based navigation for quadruped locomotion.

The rest of this letter is organized as follows. Related work is reviewed in Section II. The vision-based navigation method for quadruped locomotion is summarized in Section III. Semantic segmentation of garden scene is presented in Section IV. Section V proposes a trajectory compensation method. Section VI demonstrates the experiment results. This line of research is concluded in Section VII. 

%% file: 3-related_work.tex
\section{Related Work}

Navigation based on semantic segmentation for quadruped locomotion is yet explored. Semantic segmentation methods are mostly based on deep learning techniques and are extensively studied in autonomous driving. The dominant deep learning models include ERFNet\cite{8063438}, FCN \cite{7298965}, SegNet \cite{7803544}, etc. To improve the performance of the semantic segmentation models, The DeepLab series applies atrous convolution to magnify the receptive field without increment of weight amount, to extract large-scale context between objects \cite{7913730, 2018Encoder}. The efficiency of These models is relatively low due to the high computational cost. However, ERFNet achieves promising performance on mobile hardware. A self-supervised learning method is applied to train the semantic segmentation model with the generation of the traversable and untraversable labels with the aid of LIDAR \cite{7989025}. With the sensor fusion method, the 2D perception results is able to be projected into 3D space for obtaining the semantic map. The perception in the autonomous driving field highly relies on the big data of the urban environment. However, there lacks of open-source data for the unstructured environment, which incurs challenges for the legged robotic navigation. Moreover, the image-based semantic perception methods only provide two-dimensional results, which implies that the 3D information obtained from LIDARs or RADARs is needed to build the 3D semantic mapping. These approaches suffer from the limited payload capacity of the small target platform, such as the small-scale quadruped robot. 

\begin{figure}[t]
    \centering
    \setlength{\abovecaptionskip}{0.1 cm}
    \includegraphics[width = 2.6 in]{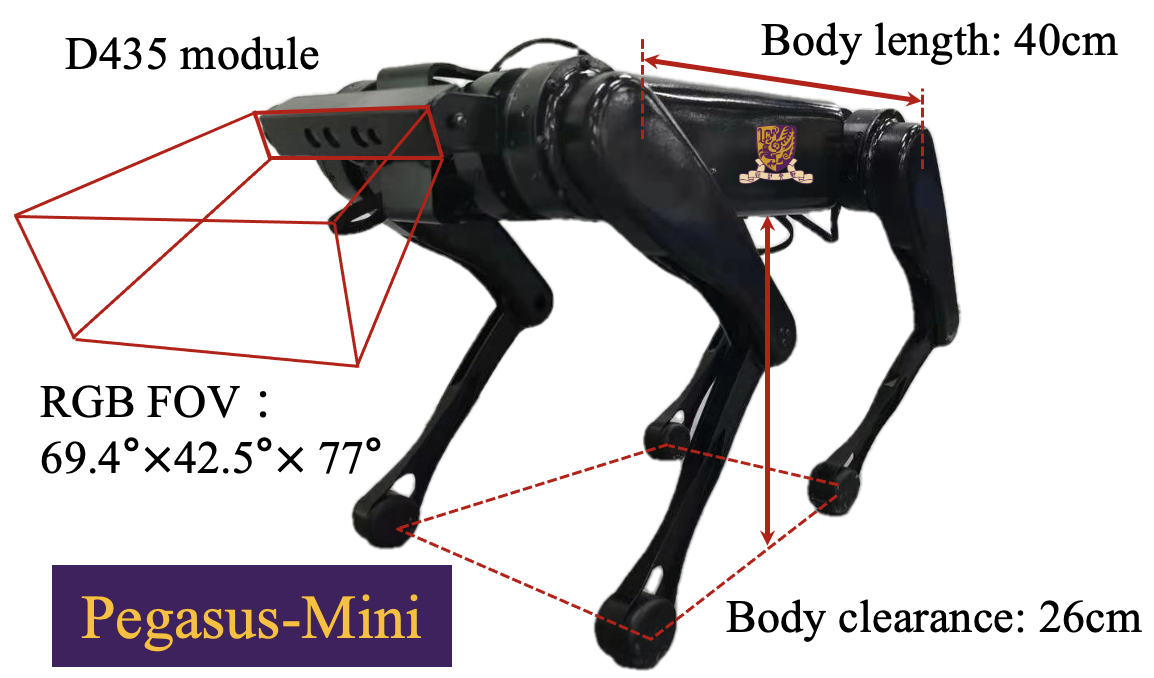}
    \caption{The small-scale quadruped robot Pegasus-Mini. The body clearance is 26 $cm$. The body length is 40 $cm$. Intel RealSense D435 camera is mounted in the front part of Pegasus-Mini to collect images for vision-based navigation.}
    \label{fig:robot}
\end{figure}

A metric of terrain negotiation difficulty is defined and a self-supervised learning-based method is developed to predict terrain properties for ANYmal locomotion \cite{8627373}. The success is impressive, however, the force-torque sensor must be equipped on the feet of the robot, which is not satisfied by many low-cost and small-scale quadruped robots. Besides, new data must be established and more workload are necessitated for the data pre-processing. Another study explores to solve the traversability estimation issue from a new perspective \cite{9247267}. In this inspiring research, a self-supervised learning method is adopted to record the acoustic signals when the robot passes various kinds of terrains and the semantic segmentation network is trained. However, this network is yet deployed on quadruped robot platforms.

In this study, a small outdoor data set is collected and labelled. The semantic feature of the terrains via only images is utilized. A convolutional neural network (CNN) of semantic segmentation is trained with the open-source dataset. To deal with the domain shift problem, a new small outdoor dataset is collected and labelled with few labors. To enhance the robustness of the path planning method via semantic segmentation method, a simple and robust algorithm is devised to compensate the robot’s pose commanded by vision-based navigation. A quadruped robot, Pegasus-Mini is built, which is able to trot at a high frequency and exhibits adaptability and flexibility to the outdoor environment. A vision-based navigation method based on semantic segmentation, together with trajectory compensation, is demonstrated on Pegasus-Mini.

%% file: 4-system_overview.tex
\section{System Overview}
\subsection{Pegasus-Mini Quadruped Robot}
The platform for the vision-based navigation test in this study is Pegasus-Mini, a small-scale quadruped robot, as shown in Fig. \ref{fig:robot}. Pegasus-Mini is electrically actuated with 12 degrees of freedom. It weighs 12 $kg$ and is 0.32 $m$ tall. A D435 camera is equipped in the front of the body. The lengths of the upper and lower leg are 0.206 $m$ and 0.228 $m$ respectively. Pegasus-Mini is able to run in a trotting gait at the height of 0.26 $m$, based on leg workspace and debugging experience.

Navigation algorithm runs on Nvidia Xavier NX, which is a low-power computer with a 6-core Carmal ARM V8.2 architecture CPU and 384 CUDA core, 48 Tensor core, 8 GB RAM. Ubuntu 18.04 with the ROS-Melodic works as the operating system. Nvidia Xavier NX is deployed for running the CNN model to segment the trail in the garden. 

Locomotion is executed on an Intel UP board low-power single-board computer with a quad-core Intel Atom CPU, 4 GB RAM. Linux with RT patch works as the operating system. UP board is used to run the low-level controller, including MPC, WBC, and state estimator.

\subsection{Framework Overview}
In this section, a vision-based navigation method for a small-scale quadruped robot Pegasus-Mini is proposed. Terrain classification through training a neural network based on the CNN framework is adopted. The training algorithm enables the deployment of a light-weight visual perception system on the small scale quadruped robot Pegasus-Mini.

\begin{figure}[t]
    \centering
    \setlength{\abovecaptionskip}{-0.1 cm}
    \includegraphics[width = 3.45 in]{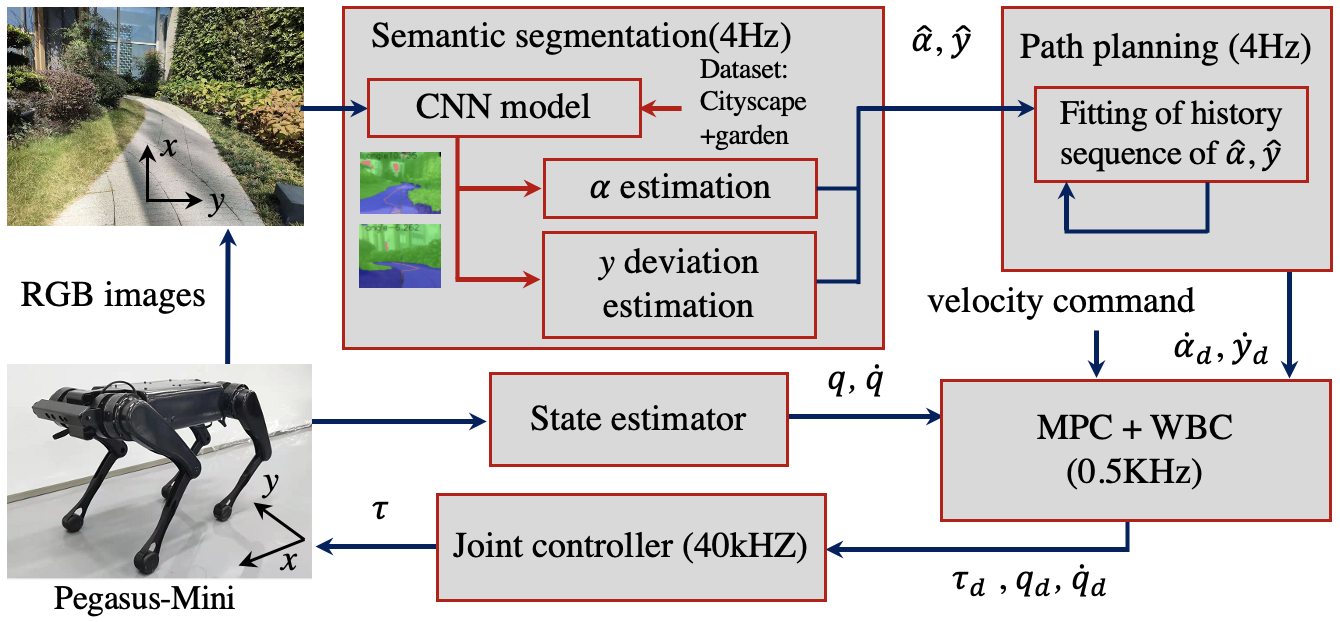}
    \caption{Vision-based navigation framework for small scale quadruped robot Pegasus-Mini.}
    \label{fig:framework}
\end{figure}

The developed framework for learning-based garden navigation is illustrated in Fig. \ref{fig:framework}. An off-the-shelf low cost camera, Intel RealSense D435 is adopted on Pegasus-Mini to collect the RGB images. Two datasets are utilized for training the CNN neural network. One dataset is collected from an open-sourced dataset and the other dataset is generated from the garden. 

The CNN model takes the RGB images sensed by the camera D435 mounted in the front part of Pegasus-Mini's body and outputs the segmentation of the traverse path. Yaw motion and $y$ deviation estimation are calculated based on the traverse path's geometrical information. Under the condition that image segmentation fails, the trajectory planning compensation method is supplemented to enhance the success rate of traversability in the garden environment. 

The path planner outputs the desired yaw angle velocity and linear velocity in $y$ direction. The desired velocity is fed into MPC and WBC to calculated the joint controller. The CNN model runs at 4 $Hz$. The path planner runs at 4 $Hz$. The MPC and WBC run at 0.5 $kHz$. The joint controller runs at 40 $kHz$.

\begin{figure}[ht]
\centering
\includegraphics[width = 3.4 in]{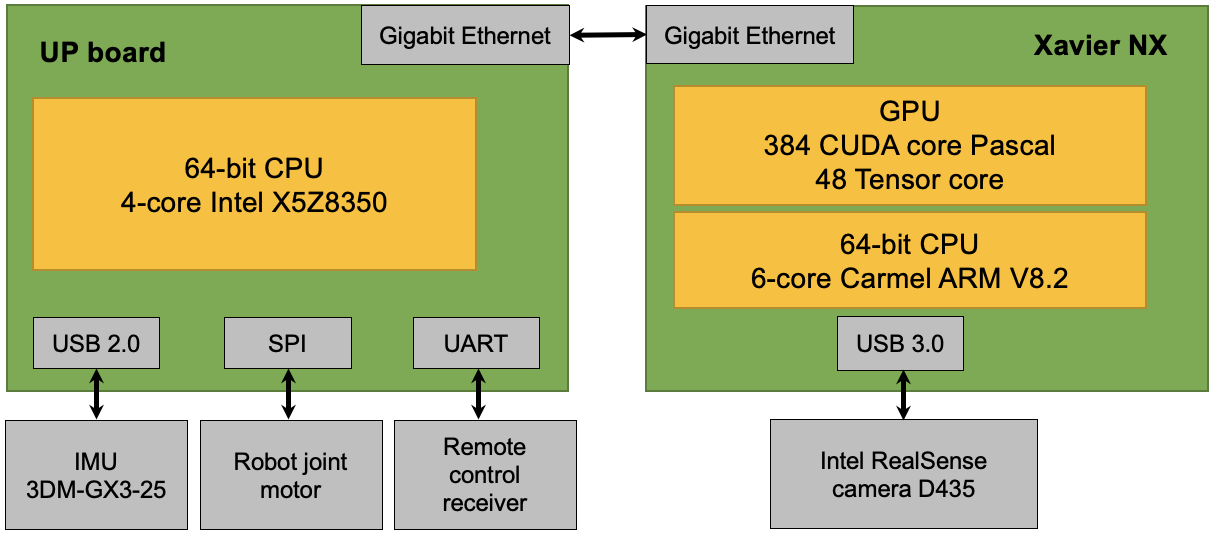}
\caption{Computing architecture of the Pegasus-Mini quadruped robot. Nvidia Xavier NX is used to run the CNN model to segment the trail in the garden. UP board is used to run the low-level controller, including MPC, WBC, and state estimator as shown in Fig. \ref{fig:framework}.}
\label{fig:computing_architecture}
\end{figure}

Fig. \ref{fig:computing_architecture} presents the computing architecture deployed on Pegasus-Mini. Nvidia Xavier NX is used to run the CNN model to segment the trail in the garden. UP board is used to run the low-level controller, including MPC, WBC, and state estimator as shown in Fig. \ref{fig:framework}. Xavier NX communicates with the UP board through ethernet. IMU communicates with the UP board through USB 2.0 to feedback the posture information. Desired joint position, velocity, and torques are calculated in UP board and sent to the robot joint controller through the SPI interface. The operator is also able to directly send command using a remote control receiver through the UART interface.

%% file: 5-semantic_segmentation.tex
\section{Semantic Segmentation of Garden Scene}
In this section, a CNN model based on ERFNet is adopted for the trail segmentation and classification in a garden. Domain adaption method, dataset, and training method will be introduced in the below subsections. The yaw motion and deviation in $y$ direction estimation based on the segmentation of trail will be also described.

\subsection{Domain Adaption and Network Training}
\begin{figure}[ht]
\centering
\includegraphics[width = 3.4 in]{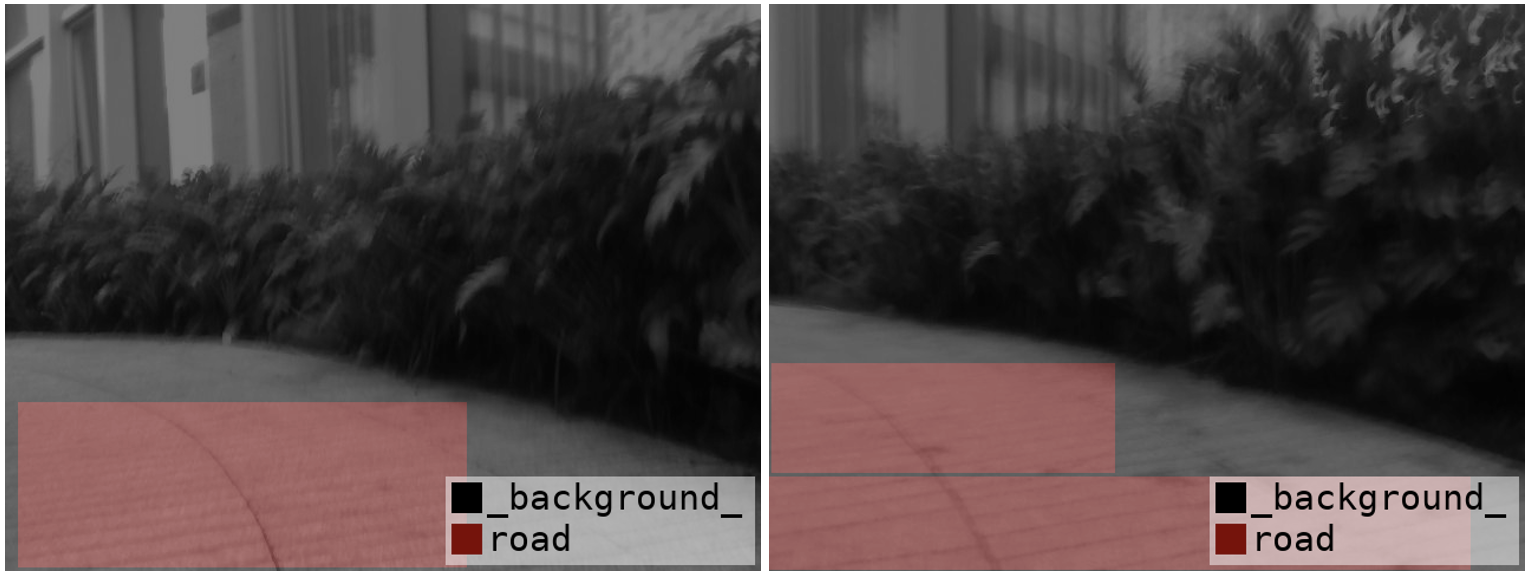}
\caption{Example of the manually collected dataset for domain adaption. Only some parts of the road class area (traversable path) are labelled with the consideration of saving human labors. The labelled area covers about 5$\%$~50$\%$ of the whole image.}
\label{fig:dataset}
\end{figure}

There exists abundant open-sourced datasets for semantic segmentation, about indoors or urban environment. However, no dataset contains only the unstructured environment. In order to solve the domain shift problem and make the Pegasus-Mini correctly perceive the garden environment, a small dataset is recorded about the garden environment. After collection, we only label the traversable area of the garden dataset with the rectangle box for two reasons. Firstly, the model needs to work in different kinds of gardens that have various backgrounds but similar path, so precise labeling with all the classes is unnecessary. Secondly, we hope to complete the labeling work with less human labors. Based on this idea we only spend an hour labeling the traversable path with the rectangle box, and the Fig. \ref{fig:dataset} shows examples of the label.

The new garden dataset is combined with the Cityscape dataset \cite{7780719} for the domain adaption. From another perspective, the Cityscape dataset provides the negative labels. To this end, the Cityscape dataset is relabelled from 30 classes into 3 classes: traversable path, untraversable path, and void. This corresponds to the garden dataset, therefore, these are able to be put together and only one-stage training is needed, to avoid the catastrophic forgetting problem during the training.

\begin{figure}[ht]
\centering
\includegraphics[width = 3.3 in]{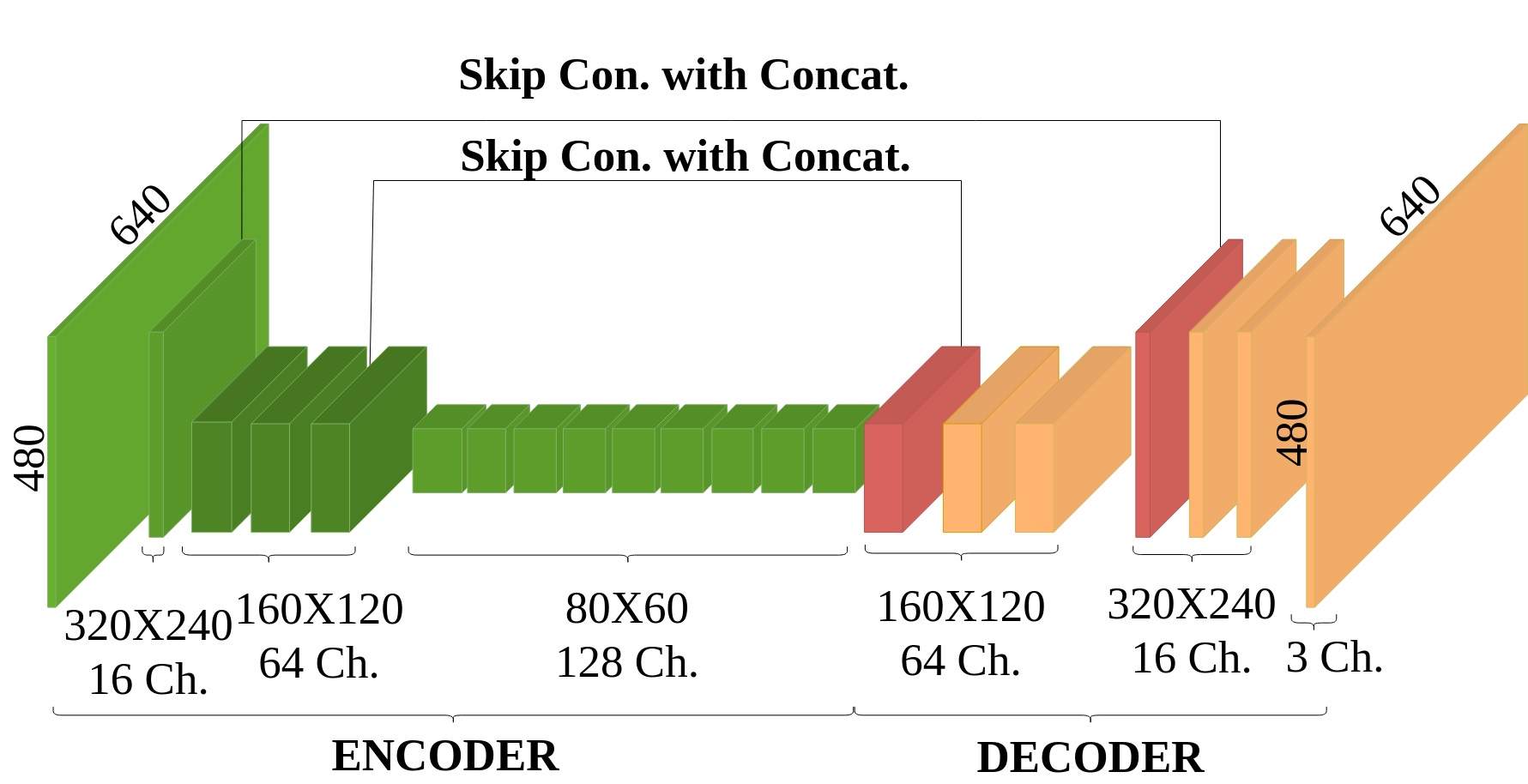}
\caption{The network architecture of CNN model adopted in this study.}
\label{fig:cnn_framework}
\end{figure}

The adopted network architecture in this study is based on ERFNet, which is capable of achieving good performance in semantic segmentation while running in real-time on mobile hardware \cite{7995966}. Compared with the original network, the size of the input layer and the number of the channels are changed in the last output layer to fit the training dataset. In order to speed up the model and reduce the over-fitting problem, the network is simplified by shrinking the repeated bottleneck from 5 to 3 and from 3 to 2 in the encoder and decoder respectively. Besides, two skip connections are introduced between the intermediate decoder and encoder layers to improve the performance of the model. The overall structure of the network model is illustrated in Fig. \ref{fig:cnn_framework}. 
Since the semantic segmentation is a pixel-wise classification task, the cross-entropy loss is employed in form of multi-class classification as (\ref{eq:1}). The dataset is reweighted so that the weight of the garden dataset and Cityscape dataset is set as 2:1, considering that the amount of garden images is much fewer than the urban images.

\begin{equation}
    J(W) = - \frac{1}{N} \sum_{i=1}^{N} \sum_{j=1}^{C} y_{i,j}log (\hat{y}_{i, j}),
    \label{eq:1}
\end{equation}
where $y_{i,j}$ and $\hat{y}_{i, j}$ are the ground truth class and predicted class of $n$ $th$ images respectively. The loss of $N$ images and $C$ classes are summarized.

\subsection{Dataset}
To train the network, two datasets are mainly used. The Cityscape semantic segmentation benchmark is a public dataset with a multi-sensor collection in Germany. The dataset includes more than 25000 annotated images of which about 5000 images are with fine annotations. The images are annotated with 30 classes of relevant objects and summarized with 8 groups. Classes like road and sidewalk can be considered as the traversable path in our case and objects like person, vegetation and terrain are classified as untraversable areas, which is nutritious for the training.

Additionally, in this study, a dataset is generated for domain adaption by controlling the robot to trot in the garden. The robot trotting speed is set at 0.7 $m/s$ with the gait frequency at 4 $Hz$, to minimize the influence of the locomotion controller. This dataset consists of only one scene, the garden, with the robot-view image recorded at 4 $Hz$. There are totally about 700 images with a resolution of 640 $\times$ 480. The first 100 images are selected as the validation sets and the rest as the training set.

\subsection{Training}
To improve the generalization performance of the model, we perform image augmentation on the input images. The images are randomly cropped from the Cityscape dataset to the resolution of 640 $\times$ 480 to fit the input. The following image pre-processing augmentations are applied to both two datasets:

1) Image normalization to [0, 1];

2) Randomly horizontal flip of the input images with probability 0.5;

3) Random rotation of images from [-5$^\circ$, 5$^\circ$].

Our model is trained using the Adam optimizer with a learning rate of $10^{-5}$. We choose the batch size of 8 and the epoch of 150. To deal with the overfitting problem, the L2 regularization and early stopping strategy are applied. An NVIDIA GeForce RTX 2080 GPU is adopted for all the training and evaluation of the models. Computation architecture is as shown in Fig. \ref{fig:computing_architecture}.

\subsection{Pose Adjustment}
The 2D perception result upstream is used for the garden navigation, to enable the robot to locomote in the middle of the path throughout. To this end, a pose adjustment algorithm is proposed for the garden navigation mission. 

The output images are downsampled to reduce the computational burden. Based on the post-processed perception results, the midpoints are calculated along the path from the boundary on both sides. The poses of all the points are summed to get the averaged point. From the start point to the averaged point, the radial angle can be calculated to provide the yaw angle for the next movement, as shown in (\ref{eq:yaw_angle}).

\begin{equation}
    \alpha = arctan \frac{ \sum_{i=1}^{N} (p_{ix} - p_{0x})}{ \sum_{i=1}^{N} (p_{iy} - p_{0y})},
    \label{eq:yaw_angle}
\end{equation}
where $p_i$ and $p_0$ are the calculated midpoints and start point respectively. $\alpha$ is the yaw angle of the robot's pose.

The overall perception and pose adjustment process can run at more than 6 $Hz$ on the robot platform after optimization. Since the trotting frequency of the robot is 4 $Hz$, the update frequency of the perception is fixed at also 4 $Hz$. In this case, the robot does not miss the perception results, and the time latency between the visual input and trotting decision is always less than 0.25 $s$, which guarantees the timely and robust movement control.

With the assistance of the classification of terrains in the garden environment, the trail is able to be segmented and extracted to provide a traversability reference for Pegasus-Mini. The edge of the trail in the image of Intel RealSense D435 is calculated for the estimation of the desired trajectory of the quadruped robot. Vision-based navigation is cost-effective and computing efficient. 4 $Hz$ frame frequency of the classification satisfies the requirement of normal locomotion speed of quadruped robot. However, there exists instability of trail classification based on the learning method.

%% file: 6-path_planning.tex
\section{Path Compensation Planning}
Section IV introduces the learning method using a neural network to segment trail in a garden environment. The method is capable of extracting the edge of the trail for estimation of the direction and the middle line of the trail. However, the learning-based method is not able to guarantee the success rate, especially when the quadruped robot happens to move to a new location where the scenario is not recognized due to the limitation of the training dataset. Pose estimation is extensively studied in legged locomotion and floating-base system \cite{luo2020estimation, luo2021modeling}. Considering the fact that the state of the quadruped robot does not jump suddenly but changes consistently, the consistency characteristic of the dynamics of the quadruped robot is utilized and a compensation method is proposed in this section to correct the issue existing in vision learning.

\begin{figure}[b]
\centering
\setlength{\abovecaptionskip}{0.0 cm}
\includegraphics[width = 3.4 in]{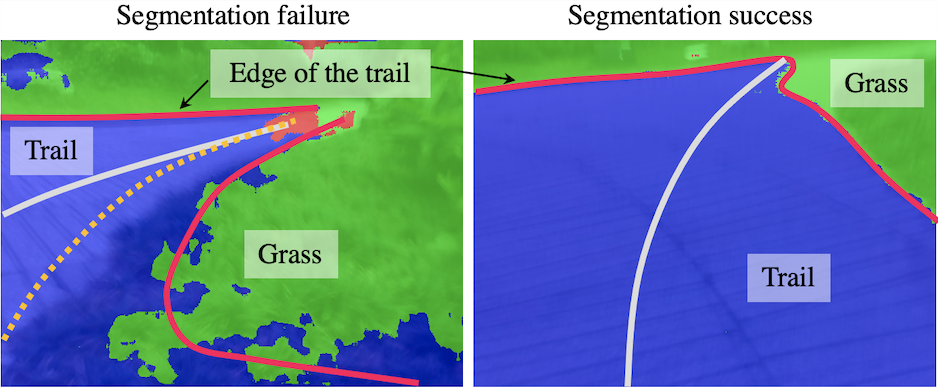}
\caption{In the left image, the grass is mistaken to be the trail and the correspondence desired yaw rotation and $y$ deviation of Pegasus-Mini are wrongly determined. In comparison, the right image provided a right desired yaw rotation and $y$ deviation.}
\label{fig:trail_comparison}
\end{figure}

As shown in Fig. \ref{fig:trail_comparison}, the trail segmentation is wrong. To compensate the mistaken trail classification, this study proposes a trajectory planning method to compensate the wrong desired yaw and $y$ deviation velocity. $y^t = [y^t_0, y^t_1, ..., y^t_{n-1}]^T$ is the vector of middle point sequence in the image collected at timestamp $t$. $y_i^t$ is the $i$ th middle point of two edges of the trail in the image. A polynomial fitting is adopted to calculate a smooth curve as the estimated path in $y$ direction.

\begin{equation}
    y^t_p = [\beta_0^t, \beta_1^t p, \beta_2^t p^2, ... , \beta_{n-1}^t p^{n-1}]^T,
\end{equation}
\label{eq:y_t}
where $y^t_p$ is the estimation of the middle point sequence in $y$ direction at time stamp $t$. $p = 0, 1, ..., n-1$. $n$ is the order of the polynomial fitting.

\begin{equation}
    y^t = P (\beta^t)^T,
\end{equation}
where $y^t = [y^t_0, y^t_1, ..., y^t_{n-1}]^T$, $\beta^t = [\beta_0^t, \beta_1^t, ... , \beta_{n-1}^t]^T$ and $P$ is:

\begin{equation}
    P = \begin{bmatrix}
    p_0 & p_1 & ... & p_{n-1}\\
    p_0^2 & p_1^2 & ... & p_{n-1}^2 \\
    ... & ... & ... & ... \\
    p_0^n & p_1^n & ... & p_{n-1}^n
    \end{bmatrix}_{n \times n}.
\end{equation}

Therefore, $\beta^t$ is able to be calculated by:

\begin{equation}
    \beta^t = (P^{-1} y^t)^T.
\end{equation}

\begin{figure}
    \centering
    \includegraphics[width = 3.4 in]{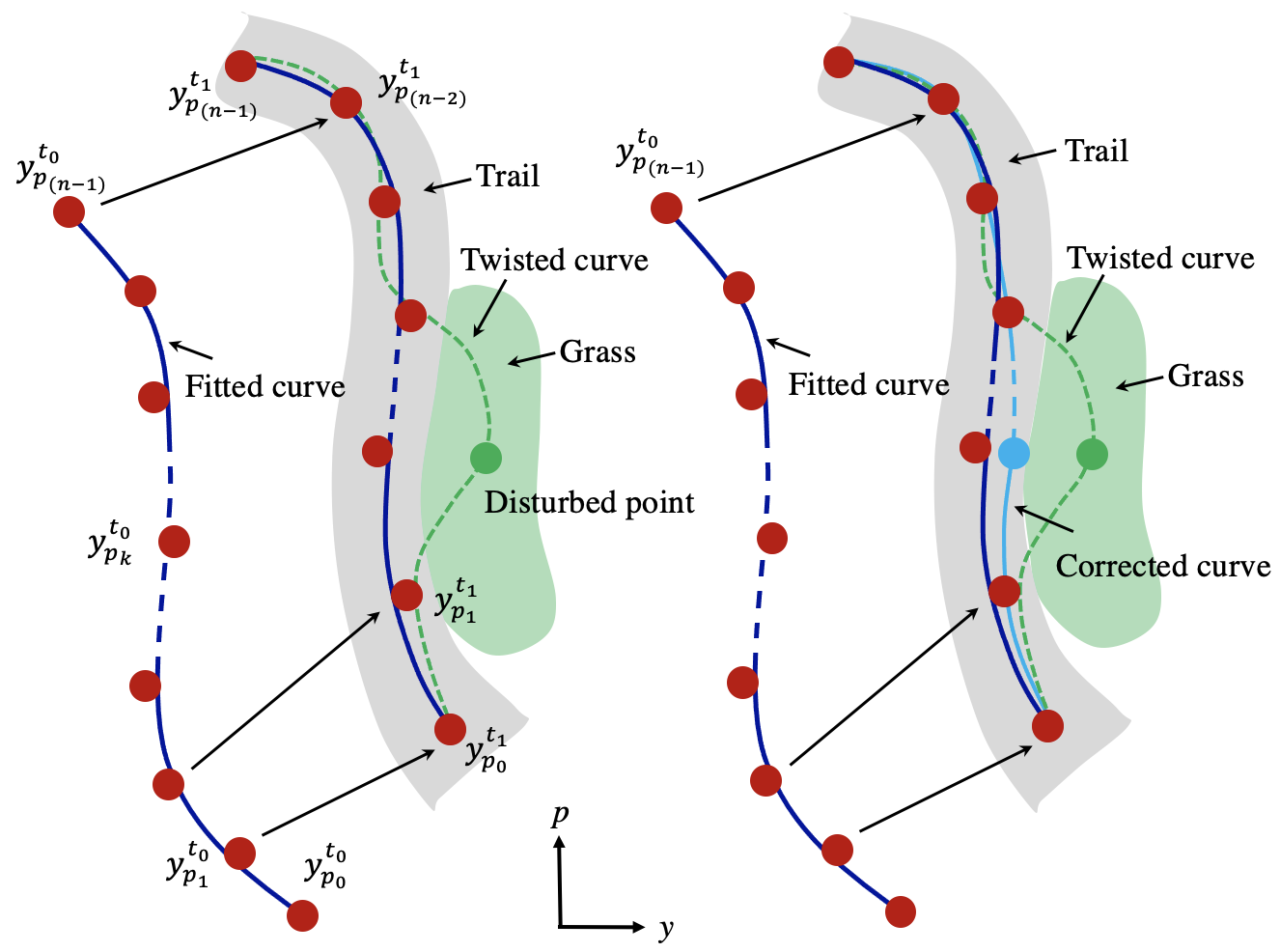}
    \caption{The polynomial fitting of the middle points estimated by CNN model. The green area corresponds to the grass terrain where Pegasus-Mini is not supposed to traverse.}
    \label{fig:fitting}
\end{figure}

Ideally, in the condition in which there is no mistaken segmentation, the coefficients $\beta^{t_j}$ at $t_{j}$ time stamp is updated with $\beta^{t_{j+1}}$:

\begin{equation}
    \beta_{p}^{t_{j+1}} = w_1 \beta_{p+1}^{t_{j+1}} + w_2 \beta_{p+1}^{t_j}
    \label{eq:iteration}
\end{equation}
where $t_j$ and $t_{j+1}$ are the $j$ th and $j+1$ th time stamp respectively. $w_1$ and $w_2$ are the weights for each term and $w_1 + w_2 = 1$. $\beta_{p}^{t_{j+1}}$ is the updated $p$ th term in the coefficient vector $\beta$ for time stamp $t_{j+1}$.

If the segmentation fails, as shown in Fig. \ref{fig:fitting}, the disturbed point calculated from the wrong trail edge (as shown in Fig. \ref{fig:trail_comparison}) will drag the fitted curve away from the desired path trajectory. In this condition, the quadruped robot will walk into the grass, which is not expected to occur.

Similarly, the estimated yaw angle of quadruped locomotion is calculated through (\ref{eq:yaw_angle}).

\begin{equation}
    \alpha_{t_j} = w_{\hat{\alpha}} \hat{\alpha}_{t_j} + w_{\alpha} \alpha,
    \label{eq:yaw_est}
\end{equation}
where $\alpha_{t_j}$ is the updated yaw angle at $t_j$ time stamp. $w_{\hat{\alpha}}$ and $w_{\alpha}$ are the weights for each term. $w_{\hat{\alpha}} + w_{\alpha} = 1$.

The principle of trajectory compensation is shown in Fig. \ref{fig:fitting}. The history information during the last $n$ th time stamp is considered together with the updated new estimation at $n+1$ th time stamp. If the estimation $\alpha$ or $y$ deviates a lot from the history records, the weight for this updated term will be attenuated.

%% file: 7-experiment.tex
\section{Experiment}
In this section, the vision-based navigation method proposed in this study is tested. The training performances of different datasets are compared and evaluated. To validate the effectiveness of the vision-based navigation on the quadruped robot Pegasus-Mini, trail detection algorithm is run under different trotting speeds from 0.2 $m/s$ to 1.0 $m/s$. Three CNN models trained using Cityscape, garden, and Cityscape-garden are compared respectively. Learn method only and the learning method with path planning compensation are compared. In the next subsections, training performance and comparison results will be demonstrated.

\subsection{Training Results}
\begin{figure}[t]
\centering
\includegraphics[width = 3.4 in]{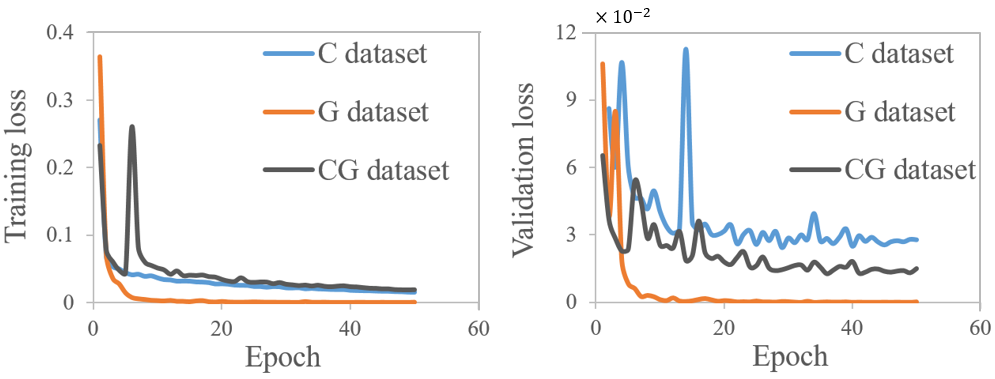}
\caption{Training loss and validation loss. The blue, orange and grey line represent the training and validation loss results based on models of Cityscape ($C$ dataset), garden ($G$ dataset) and Cityscape-garden ($CG$ dataset) respectively.}
\label{fig:training_results}
\end{figure}

Training and validation loss results are as shown in Fig. \ref{fig:training_results}. In Fig. \ref{fig:training_results}, the blue line represents the training loss of the Cityscape dataset, the orange line represents the training loss of the garden dataset and the green line represents the training loss of the Cityscape and garden dataset. From Fig. \ref{fig:training_results}, it is demonstrated that the training performance based on three datasets is satisfactory. 

\begin{figure}[ht]
\centering
\includegraphics[width = 3.0 in]{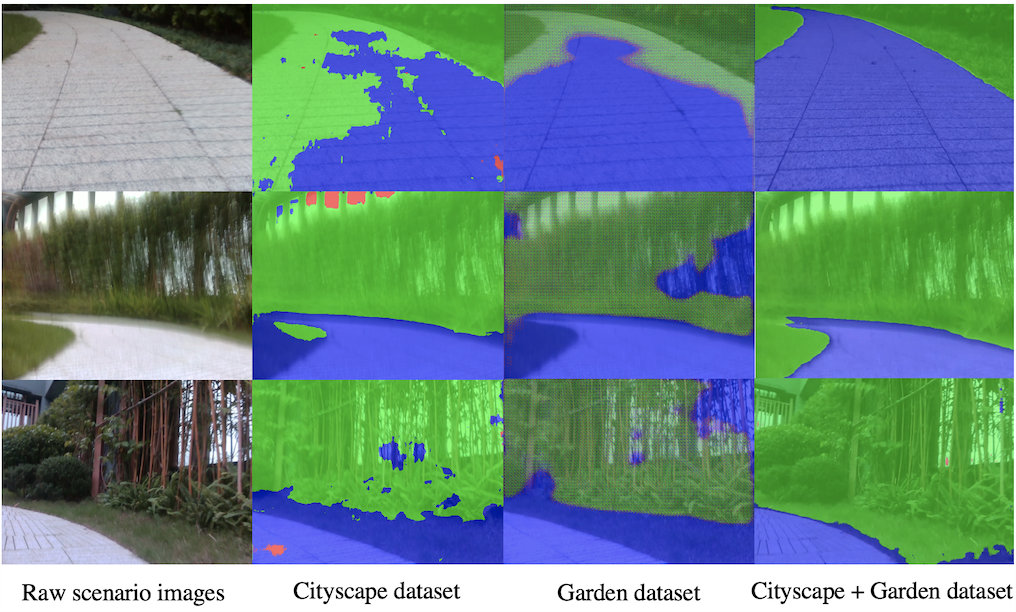}
\caption{Qualitative prediction results of semantic segmentation trained with difference dataset. The blue, green and red regions represent the traversable area(road), untraversable area and sky respectively.}
\label{fig:prediction}
\end{figure}

In this test, the CNN models trained with three datasets are tested on the quadruped robot Pegasus-Mini. As shown in Fig. \ref{fig:prediction}, the left column represents the raw scenario images. The left second, third and fourth columns represent the segmentation results using datasets from Cityscape, garden, and Cityscape-garden respectively. For convenience, the models trained with Cityscape, garden, and Cityscape-garden datasets are termed as $C$, $G$, and $CG$ respectively. 

\subsection{Path Planning With Compensation}
This subsection demonstrates the comparison results. The trail segmentation results using three datasets are compared under different trotting speeds. The vision-based navigation with learning method only and with the trajectory compensation are also compared. To test the effectiveness of the proposed vision-based navigation on quadruped locomotion, experiments are conducted in trotting gait at different speeds ranging from 0.4 to 1.0 $m/s$ which are common speeds for quadruped locomotion.

\begin{figure}[ht]
\centering
    \setlength{\abovecaptionskip}{-0.1 cm}
    \setlength{\belowcaptionskip}{-1 cm}
\includegraphics[width = 3.0 in]{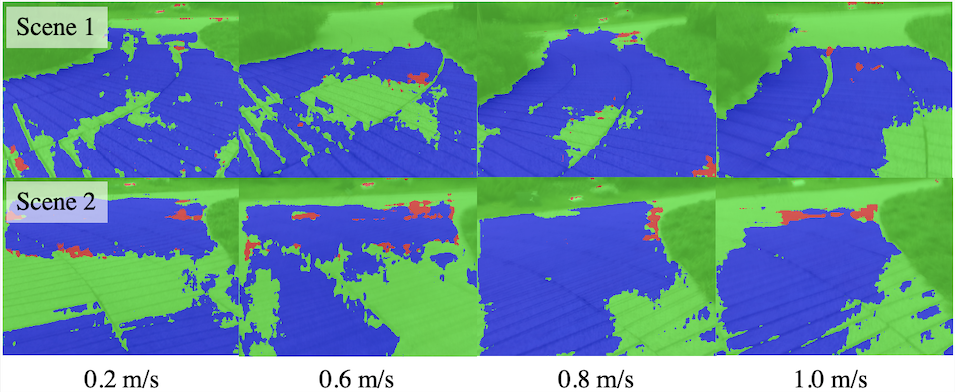}
\caption{The performance of trail segmentation using Cityscape dataset under trotting speed at 0.2, 0.4, 0.6 0.8 $m/s$ respectively.}
\label{fig:com_cityscape}
\end{figure}
\vspace{-4 mm}
\begin{figure}[ht]
\centering
    \setlength{\abovecaptionskip}{-0.1 cm}
    \setlength{\belowcaptionskip}{-1 cm}
\includegraphics[width = 3.0 in]{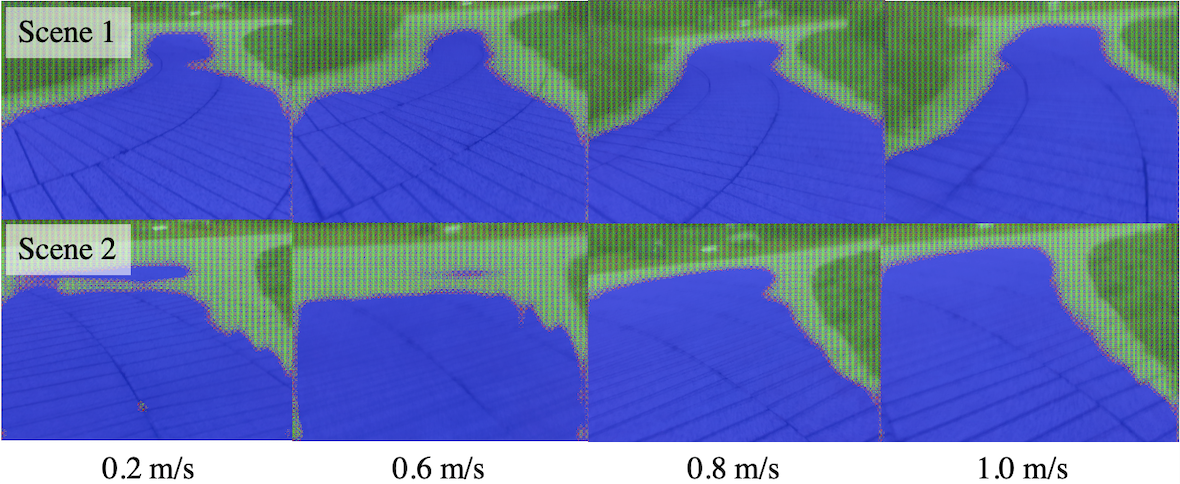}
\caption{The performance of trail segmentation using garden dataset under trotting speed at 0.2, 0.4, 0.6, and 0.8 $m/s$ respectively.}
\label{fig:com_garden}
\end{figure}
\vspace{-4 mm}
\begin{figure}[ht]
\centering
    \setlength{\abovecaptionskip}{-0.1 cm}
    \setlength{\belowcaptionskip}{-1 cm}
\includegraphics[width = 3.0 in]{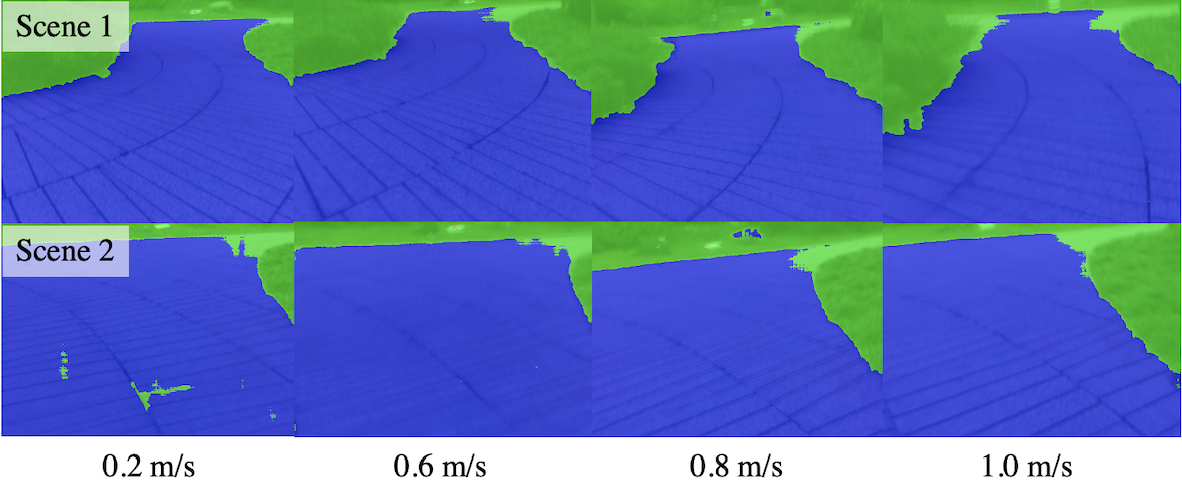}
\caption{The performance of trail segmentation using Cityscape-garden dataset under trotting speed at 0.2, 0.4, 0.6, and 0.8 $m/s$ respectively.}
\label{fig:com_cityscape_garden}
\end{figure}

\subsubsection{Results Under Different Trotting Speeds and CNN models}
In this study, a common gait, trotting, is selected to test the effectiveness of the deployment of vision-based navigation for quadruped locomotion. Four trotting speeds are set in the experiment, i.e. 0.2, 0.4, 0.6, and 0.8 $m/s$. CNN models trained with Cityscape and Cityscape-garden datasets are tested respectively.

Fig. \ref{fig:com_cityscape} shows two example images taken with $C$ model. Each row shows the trail segmentation results under four different trotting speeds. The segmentation results at four different trotting speeds do not provide a clear extraction of the trail. Overall performance is not very satisfactory.

Fig. \ref{fig:com_garden} shows the results taken with $G$ model under four trotting speeds. Each row shows the trail segmentation results in a certain scene under four different trotting speeds. Due to the small size of training the dataset of the garden, the visualized segmented images in Fig. \ref{fig:com_garden} is sparse.

\begin{figure}[ht]
\centering
\includegraphics[width = 3.0 in]{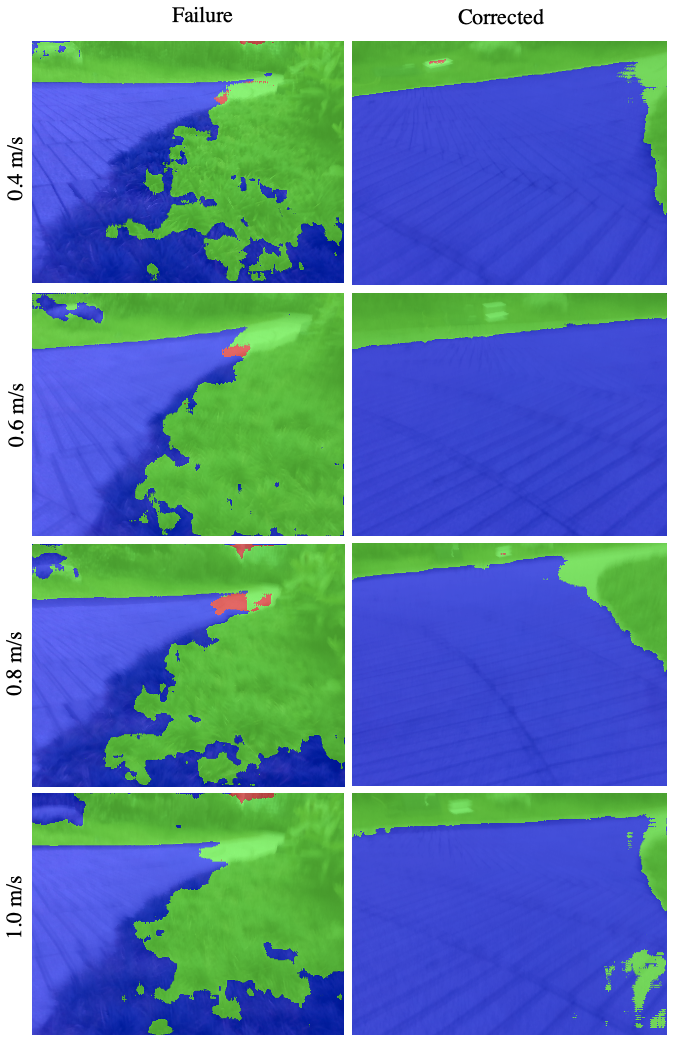}
\caption{Vision-based navigation with and without trajectory compensation under four different speeds ranging from 0.4 to 1.0 $m/s$. The left column represents the failure of the segmentation of trail from surrounding grass. The right column represents the successful corrected segmentation when trajectory compensation is supplemented.}
\label{fig:_with_compensation_comparison}
\end{figure}

Fig. \ref{fig:com_cityscape_garden} shows results taken with $CG$ model under four trotting speeds. Each row shows the trail segmentation results under four different trotting speeds. Compared with the segmentation performances in Fig. \ref{fig:com_cityscape} and \ref{fig:com_garden} which correspond to the training results using $C$ and $G$ , the $CG$ is able to output better segmentation of trail, which is taken as the traversable path for the quadruped locomotion.

\begin{figure*}[ht]
\centering
\includegraphics[width = 6.9 in]{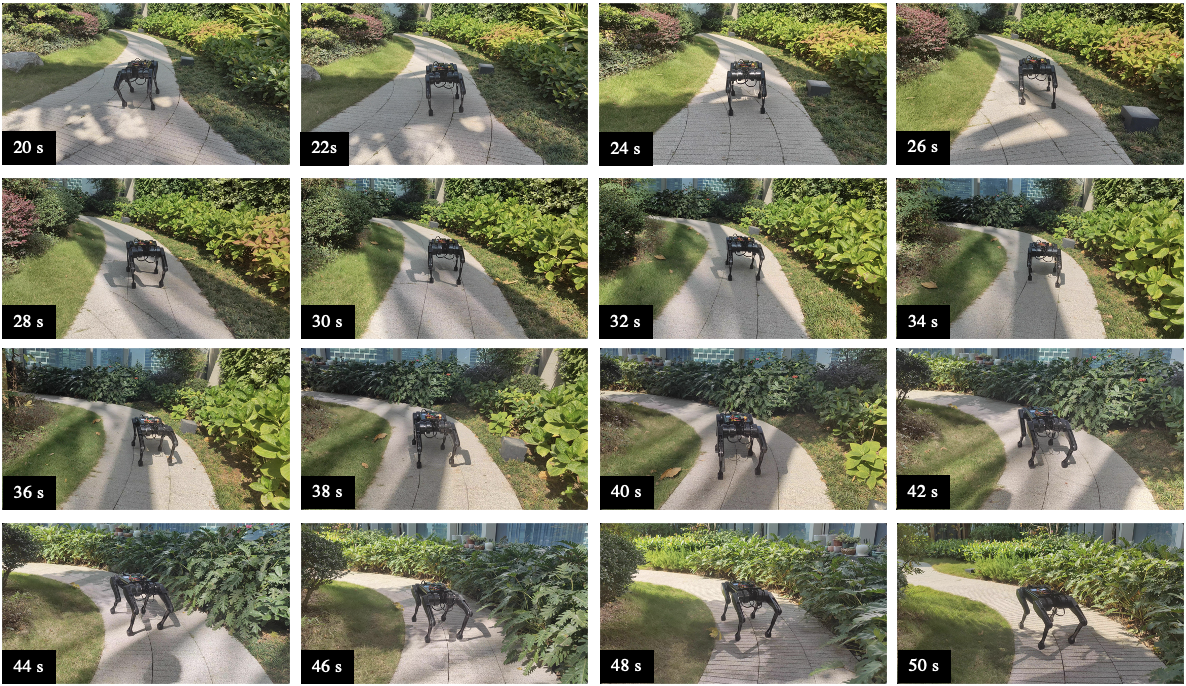}
\caption{The screenshots of the vision-based navigation of Pegasus-Mini in the garden. The time duration between each screenshot is 2 $s$.}
\label{fig:screenshot}
\end{figure*}

\subsubsection{Results of Trajectory Compensation}
In this subsection, vision-based navigation performances with and without trajectory compensation are compared. The comparison is demonstrated under four different speeds. Fig. \ref{fig:_with_compensation_comparison} shows the result under trotting speed of 0.4, 0.6, 0.8, and 1.0 $m/s$ respectively. The left column represents the wrong segmentation of the trail no matter what speed the quadruped robot trots at. In comparison, the right column shows the corrected trail segmentation when trajectory compensation is supplemented in the whole navigation framework.

The above experiment results demonstrate that the proposed vision-based navigation method is effective for the normal quadruped trotting gait in a garden environment. Despite the instability of the image segmentation for the path planning, a compensation method is supplemented to enhance the success rate of traversability in the garden environment. The screenshot of the vision-based navigation of our small-scale quadruped robot Pegasus-Mini is as shown in Fig. \ref{fig:screenshot}. It is noteworthy that the open-sourced dataset Cityscape is proven to be able to be generalized to the garden scene in this study. With a small dataset collected from a specific scene merged with the Cityscape dataset, the CNN model is able to be deployed in the field.

%% file: 8-conclusion.tex
\section{Conclusion}
This study proposed a vision-based navigation method combining learning-based method and trajectory planning to enhance the traversability. The learning method is based on ERFNet which is extensively used for semantic segmentation. The open-sourced dataset Cityscape is combined with the dataset collected from our garden scene to train ERFNet. ERFNet is deployed on a small-scale quadruped robot Pegasus-Mini to accomplish the real-time terrain segmentation. The training performance is compared with Cityscape only and with Cityscape-garden. Cityscape model ($C$ model) and Cityscape-garden model ($CG$ model) are tested in a common quadruped gait, trotting, under different speeds ranging from 0.4 to 1.0 $m/s$. Test results demonstrate that the $CG$ model performs better in the trail segmentation in the garden scene. Different trotting speeds ranging from 0.4 to 1.0 $m/s$ have little disturbance to the images sensing. However, the CNN model is not able to guarantee the stable trail extraction. To tackle this issue, this study proposes a trajectory compensation method, in which the consistent history trajectory sequence is taken into account together with the updated estimation of the middle line of the trail and the yaw angle. The learning-based method for image processing and semantic segmentation combined with the trajectory compensation method is capable of increasing the success rate of traversability in a garden scene. The future work includes further increasing the success rate traversability of quadruped locomotion.